\documentclass[11pt,final]{clv3}
\usepackage{hyperref}
\usepackage{times}
\usepackage{latexsym}
\usepackage{textcomp}

\usepackage{microtype}

\hypersetup{colorlinks=true,citecolor=blue, linkcolor=blue, urlcolor=blue, breaklinks=true}

\title{Disembodied Machine Learning: On the Illusion of Objectivity in NLP}

\author{Zeerak Waseem}
\affil{University of Sheffield}

\author{Smarika Lulz}
\affil{Humboldt University, Berlin}
\author{Joachim Bingel}
\affil{Hero I/S}
\author{Isabelle Augenstein}
\affil{University of Copenhagen}

\date{}

\begin{document}
\maketitle
\runningtitle{Disembodied Machine Learning}
\runningauthor{Waseem, Lulz, Bingel \& Augenstein}
\begin{abstract}
Machine Learning seeks to identify and encode bodies of knowledge within provided datasets. However, data encodes subjective content, which determines the possible outcomes of the models trained on it. Because such subjectivity enables marginalisation of parts of society, it is termed (social) `bias' and sought to be removed. In this paper, we contextualise this discourse of bias in the ML community against the subjective choices in the development process. Through a consideration of how choices in data and model development construct subjectivity, or biases that are represented in a model, we argue that addressing and mitigating biases is near-impossible. This is because both data and ML models are objects for which meaning is made in each step of the development pipeline, from data selection over annotation to model training and analysis. Accordingly, we find the prevalent discourse of bias limiting in its ability to address social marginalisation. We recommend to be conscientious of this
, and to accept that de-biasing methods only correct for a fraction of biases.
\end{abstract}
\section{Introduction}
Machine Learning (ML) is concerned with making decisions based on discernible patterns observed in data. Frequently, ML models and the bodies of data they act on are divorced from the context within which they are created, leading to an imposed `objectivity' to these processes and their results. Given that supervised ML seeks to distinguish a set of given bodies of data from one another, and unsupervised ML aims to identify discernible bodies of data in the data provided;\footnote{Bodies of data are amalgamated entities that exist by virtue of a strict separation from the material bodies they are derived from.} both the underlying data and the model applied to it strongly influence what bodies are discovered, and what may be discovered within these bodies. ML models were initially hailed as objective, unimpeded by subjective human biases, and by extension by social marginalisation \cite{ONeil:2016}. However, more and more research suggests that social biases are common in ML models, and that such biases in the underlying data may be exacerbated by the ML models \cite{Zhao:2017}. Accordingly, a number of research directions seek to identify \cite{Shah:2020, Bender-Friedman:2018, Mitchell:2019, Buolamwini-Gebru:2018}, reduce or remove social biases \cite{Zhao:2017, Agarwal:2018} from ML models to protect against further marginalisation. 
However, previous work frequently assumes a positivist logic of social bias as an optimisation problem, i.e.~that bias is a finite resource can be disentangled, isolated, and thus optimised for.

We revisit these assumptions and question solutionist approaches that dominate the ML 
literature.
Drawing on work from feminist Science and Technology Studies (STS) \cite{Haraway:1988} and examples from Natural Language Processing (NLP), we argue that: (a) bias and subjectivity in ML are inescapable and thus cannot simply be removed; therefore (b) 
requires an ongoing reflection on the positions and the imaginary objectivity that ML researchers and practitioners find in subjective realities reflect political choices in the ML pipeline. 
By contextualising bias in these terms, we seek to shift the discourse away from bias and its elimination towards subjective positionality.

\section{Previous Work}
Previous work on bias in ML: (i) maps models and datasets with their intended uses and limitations; (ii) quantifies and analyses disparities; or (iii) mitigates biases present in models and datasets.

\paragraph{Mapping}
\namecite{Bender-Friedman:2018} propose `data statements', a tool to describe and expose representational biases in the processes of developing datasets from collection to annotation. Analogously, \namecite{Mitchell:2019} propose `model cards' to describe ML models and their behaviour across different populations that might be subject to a given model along with its intended use. Similarly drawing on \namecite{Haraway:1988}, \namecite{Rettberg:2020} identifies how data is situated and viewed through disembodied positions in the aggregation and display of personal data in mobile applications.

\paragraph{Quantification} \namecite{Shah:2020} propose a mathematical framework quantifying biases in different steps in the NLP pipeline, basing their conceptualisation on work on ethical risks for NLP systems by \namecite{Hovy-Spruit:2016}. More practically, \namecite{Buolamwini-Gebru:2018} identify how commercial facial recognition systems perform and fail for people with darker skin and women, and perform worst for women with dark skin. Turning to language, \namecite{Gonen:2019} highlight that methods for debiasing word-embeddings leave traces that allow for reconstructing gendered spaces in `debiased' word embeddings.

\paragraph{Mitigation} 
Two conceptualisations of bias can be found in the large body of work on addressing model biases \cite[e.g.][]{Agarwal:2018,Romanov:2019,Kulynych:2020,Bolukbasi:2016}: one in which bias is imagined as a finite quantity in the model that can be minimised by altering the model's representation \cite{Agarwal:2018,Romanov:2019};\footnote{This line of work has the dual aims of minimising discrimination, while maximising performance for a given metric.} and one which, similar to our work, accepts the premise that ML, and more broadly optimisation systems, contain social biases \cite{Kulynych:2020}. Working with the latter assumption, \namecite{Kulynych:2020} propose a class of systems that use optimisations logic to counteract the marginalisation a group experiences as the result of ML being applied to them.

\section{The `God Trick' of Objectivity}
In her seminal STS work, Donna Haraway \citeyearpar{Haraway:1988} calls into question the notion of objectivity, arguing that the production of knowledge is an \textit{active} process, in which we subjectively construct knowledge based on our very particular, subjective bodies. She argues that an `objective' position like all other positions comes with its own limitations in what it obscures and highlights. In other words, an `objective' position is no less subjective, insofar it privileges the point of view of a particular body marked by subjective social and political meanings and possibilities along lines of race, class, geography,  gender etc. However, unlike other `subjective' positions, an `objective' position claims omniscience for itself by denying its particular embodiment, thereby obscuring its own subjective rootedness. This position can then be understood as a disembodied subjective position. By denying the subjectivity of its own body, the objective position elevates itself over other `lesser subjective bodies', thus playing the `God trick' \cite{Haraway:1988}.

Through its disembodiment, the position of objectivity claims to be `universal' and free from embodied socio-political meaning and is therefore applicable in all contexts and can thus be imposed upon all other subjective positions \cite{Mohanty:1984}. Consequently, embodied positions are mired in a particular (as opposed to `universal') context and their particularised experiences of embodied positions can safely be rejected, as accepting them would threaten the omniscient claim of objective study. 
However, as \namecite{Haraway:1988} argues, subjectively embodied positions allow for things to be made visible, that are otherwise invisible from the disembodied position. For instance, in the context of \textit{n-word} usage, an exclusive focus on its derogatory use would imply understanding the word through a disembodied and universalised position, which is a position often (but not always) occupied by the white human body in our world. It is only through an engagement with the particularised experiences of black bodies that the rich cultural meaning crafted in African-American communities reveal themselves \cite{Rahman:2012}.

\section{Embodiment in the ML Pipeline}
Haraway's \citeyearpar{Haraway:1988} critique of objectivity makes it possible to understand subjectivity or bias in ML in a way that recognises its potential to create social marginalisation, without inherently reducing it to a problem which can be optimised. We argue that in ML, the disembodied or objective position exists: (i) in the person designing the experiment and pipeline by developing methods to apply to a dataset of \textit{others}; (ii) in the data which is often disembodied and removed from context, and potentially given adjudication by externalised others that may not be aware of the final use of their work; and (iii) in the model trained on the embodied data subjects.\footnote{We highlight here the inherent self-contradiction in ML taking the position of objectivity while tacitly accepting that it is subject to disembodied data as evidenced by the fields of domain adaptation and transfer-learning.} 

We note that once data are ready to be processed by the model, we can consider the model to embody the data, as it is limited to the bodies of knowledge it is presented with. Thus, all other positions, i.e. those not represented in the training data, become disembodied. This can help explain why ML practitioners frequently call for `more' and `more diverse' data \cite{Holstein:2019} to address models that are unjust. However, simply adding more data without addressing whom the datasets embody and how is unlikely to yield the desired result of more just and equitable models.

\paragraph{Embodiment of the designer} A lack of diversity in ML teams is often attributed as a source of socially biased technologies with corresponding calls for increasing embodying diverse experiences \cite{West:2019}. The embodied designers, through data and modelling choices, project an embodiment of self into the technologies they develop.
Considering \namecite{Haraway:1988}, it is only through the recognition of different embodiments and promoting them that certain perspectives, understandings, and uses can be achieved. Thus diverse representation in designers in a team may aid in highlighting discriminatory outcomes of machine learning systems, it does not foster questions of subjective positioning giving this explicit attention.

\subsection{Embodiment in Data}

Datasets, following \namecite{Haraway:1988}, can be understood as a form of knowledge that does not simply exist but is produced \cite{Gitelman:2013} through embodied experiences. Subjectivity can stem from various sources, including the data source \cite{Gitelman-Jackson:2013}, the sampling method \cite{Shah:2020}, the annotation guidelines \cite{Sap:2019}, and the annotator selection process \cite{Waseem:2016,Derczynski:2016}. 

We ground our discussion of how subjectivity manifests itself in ML through processes of meaning-making, modelling choices, and data idiosyncrasies. 
A common denominator we seek to highlight is the subjective and embodied nature of data and subsequent classifications; that by taking a position of objectivity, one cannot do justice to the needs 
of individual or 
discernible communities.

\paragraph{High-level tasks} 
A range of NLP tasks are highly sensitive to subjective values encoded in the data. This includes high-level tasks that require semantic and pragmatic understanding, e.g. machine translation (MT), dialogue systems, metaphor detection, and sarcasm detection. In MT, research has identified a range of issues, including stylistic \cite{Hovy:2020} and gender bias \cite{vanmassenhove2018getting}. 

Issues pertaining to the reinforcement of sexist stereotypes have been the object of academic and public scrutiny. A classic example is the stereotypical translation of English \textit{doctor} (unmarked for gender) to German  \textit{Arzt} (marked for masculine), while \textit{nurse} (unmarked) is translated to \textit{Krankenschwester} (feminine). Here, the `objective' position is a patriarchal one, which delegates more prestige to men and less to women. The translations above may be correct in certain, but not all contexts.
This exemplifies the overarching problem that there is rarely one single `gold' label for a given document \cite{Reiter:2018}, yet most training and evaluation algorithms assume just that.

In text simplification, numerous datasets postulate that some words, sentences or texts are difficult, while others are simple. These labels are typically provided by human annotators, and while there might be clear majorities for the labelling of certain items, the disembodied position and generalisational power of the annotations will never do justice to the subjective embodiments of text difficulty both across user groups (language learners of different L1 backgrounds, dyslexics, etc.) and just as much within these groups.\footnote{There is some merit in the meta-information on 
task-relevant demographic variables of individual annotators 
in the datasets for the Complex Word Identification 2018 Shared Task. Further, recent work recognises that text simplification systems must build on personalised models \cite{yimam2018par4sim, lee2018personalizing,bingel2018lexi}.}

For abusive language detection, the causes and effects of embodiment in different stages have been considered in a dataset for offensive language use \cite{Davidson:2017}. \namecite{Waseem:2018} argue that a consequence of embodying a white perspective of respectability is that almost all instances of the \textit{n-word} are tagged as the positive classes. \namecite{Sap:2019} show that by indicating the likely race\footnote{As assumed through the prediction of dialect.} to the annotators, they seek to align their embodiment of `offensive' with the author's dialect. Further, \namecite{Davidson:2019} argue that the initially sampled data may itself contain social biases due to a disembodied perspective on slurs.

\paragraph{Core NLP tasks} However, the issues outlined above are far from limited to high-level NLP tasks. Even core tasks such as part-of-speech (POS) tagging are sensitive to the subjective nature of choices in the ML pipeline. Consider the Penn Treebank tagset \cite{PTB:1993}, the \textit{de-facto} standard for describing English word classes. Behind this collectively accepted `objective' truth is a linguistic theory that licenses a certain set of POS tags while not recognising others. The theory, in turn, is subjective in nature, and typically informed by observations on specific kinds of language. The tagset is thus better suited to describe the kind of English its underlying theory was built on rather than other varieties, sociolects or slang. This becomes more drastically apparent when a tagset developed for English is, for better or worse, forced upon some other languages \cite{Tommasel:2018}.

\subsection{Embodiment in Modelling}
While datasets are a large source of how a model may be embodied, ML models also encode which positions, or embodiments, are highlighted. 
Model behaviour can be seen as being on a spectrum ranging from globally acting models, i.e.~models that compound multiple senses of word usage with little regard to its local context; and locally acting models, which seek to embody the datum in the context it is created in, e.g. context-aware models \cite{Garcia:2019,Devlin:2019}.


By virtue of the subjective nature of grounding datum in context, there is a large variation in how locally acting models may be developed. Transfer learning can provide one possible avenue for locally acting models. Through transfer learning, knowledge produced outside of the target task training set can alter what a model embodies. For instance, should a dataset embody the language production in multiple sociolects, a pre-trained language model \cite{Devlin:2019}\footnote{Similar issues affect contextual models \cite{Tan:2019} as sociolects and dialects may not be well represented in their training data \cite{Dunn:2020}.} or mixed-member language models \cite{Blodgett:2016} may provide deeper information about the sociolects in question by examining the sentential context.\footnote{While `context' here refers to sentential context, 
language production is situated within a larger socio-political context.} It is important to note that the large-scale datasets for language models  rely on disembodying the data from the bodies creating them to identify collective embodiments.
Similarly, multi-task learning models can offer a path to embodying the creator of the datum through author attribute prediction as auxiliary task(s) \cite{Benton:2017,Garcia:2019}, thus allowing models to take into account the embodiment of the datum. 

\section{Discussion}
If subjective choices or biases masquerading as disembodied `objective' positions permeate through the ML pipeline -- and we argue that they do -- the quest for objectivity or bias-free ML becomes redundant. Rather, such a quest for objectivity or a universal `truth' may further harm already marginalised social groups by obscuring the dominance of certain bodies over others. Any effort to obscure only deepens the power of dominant groups and hurts marginalised communities further by justifying the imposition of experiences of dominant bodies upon marginalised bodies under the guise of `objective' or `bias-free'.

Designers of ML models and pipelines become complicit in how these marginalise when they fail to recognise their own positionality.
Through a recognition of one's embodiment, designers can account for what (and whom) their position and models derived from it, allow and penalise, and the political consequences thereof.
%
As data permeate the ML pipeline, a consideration of how data is embodied can allow for answering specific questions embodied in context; that the contexts which create data are present in every step of the dataset creation pipeline; and that as contexts change, so does the applicability of data.
Further, models themselves privilege some views over others, and while transfer learning provides some avenues for embodying data in the model, what positions are given space remains a political question.

The discourse on bias in ML does look to account for these political consequences. However, it pins the problem down to the presence of subjective, embodied or ``biased'' positions in ML models and seeks their eradication. 
Thus, we propose to let go of fairness as a matter of bias elimination in a solutionist endeavour without regard for subjective experiences. 
Shifting to consider embodiments would instead require one to reflect on the subjective experiences that are given voice
, as well as which bodies one needs to account for to give voice to socially marginalised groups.

\bibliography{eacl2021}
\bibliographystyle{compling}
\end{document}